%% file: root.tex
\documentclass[letterpaper, 10 pt, conference]{ieeeconf}%

\IEEEoverridecommandlockouts%

\usepackage{multirow}
\usepackage[percent]{overpic}
\usepackage{graphicx}
\usepackage{amsfonts}
\usepackage{soul}

\usepackage{amsmath}
\usepackage{amssymb}
\usepackage{booktabs}
\usepackage{pifont}%

\newcommand{\cmark}{\textcolor{green!60!black}{\ding{51}}}%
\newcommand{\xmark}{\textcolor{red!70!black}{\ding{55}}}%

\usepackage{cite}%
\usepackage[hidelinks]{hyperref}%
\usepackage[capitalise,nameinlink]{cleveref}
\PassOptionsToPackage{dvipsnames,table}{xcolor}%
\usepackage{xcolor}

\crefname{figure}{Fig.}{Figs.}
\Crefname{figure}{Figure}{Figures}
\crefname{table}{Table}{Tables}
\crefname{equation}{Eq.}{Eqs.}
\crefname{section}{Sec.}{Secs.}
\crefname{subsection}{Sec.}{Secs.}
\crefname{algorithm}{Alg.}{Algs.}

\title{\LARGE \bf
WideDepth: Millimeter-Accurate Benchmark for \\Fisheye Depth Estimation
}

\author{Ilia Indyk$^{1*}$, Ignat Penshin$^{1}$, Ivan Sosin$^{1}$, Maxim Monastyrny$^{1}$, Aleksei Valenkov$^{1}$ and Ilya Makarov$^{2,3}$%
\thanks{$^{*}$Corresponding author: Ilia Indyk (ilya.indyk@gmail.com)}
\thanks{$^{1}$%
Robotics Center, Moscow, Russia
        }%
\thanks{$^{2}$%
AXXX, Moscow, Russia%
        }%
\thanks{$^{3}$%
Trusted AI Research Center, RAS, Moscow, Russia}
}

\begin{document}

\maketitle
\thispagestyle{empty}
\pagestyle{empty}

\begin{abstract}
Fisheye cameras are increasingly adopted in robotics for near-field manipulation, navigation, and immersive perception, yet indoor depth benchmarks with accurate ground truth are still missing. To address this, we introduce \textbf{WideDepth} — the first indoor dataset for fisheye depth estimation, featuring 101 scenes containing 5K high-resolution stereo pairs labeled with millimeter-level ground truth depth and disparity. Our dataset also includes paired pinhole and fisheye samples across varying fields of view and baselines in both horizontal and vertical stereo setups. We further propose a method to adapt pinhole-trained stereo models to fisheye images and introduce a novel stereo fisheye image generation pipeline based on high-resolution LiDAR scans. Leveraging these methods, we thoroughly evaluate state-of-the-art monocular depth, stereo matching, and depth completion models on our benchmark. Additionally, we provide 18K LiDAR-derived sparse depth training samples, achieving up to a 62\% performance boost on fisheye data when fine-tuning pinhole-based stereo models. In summary, the high precision and versatility of our benchmark set a strong foundation for advancing research in fisheye depth estimation and robotics perception.

Project page: \href{https://ilyaind.github.io/WideDepth}{ilyaind.github.io/WideDepth}

\end{abstract}

\input{1_intro}
\input{2_related_work}
\input{3_projection}
\input{4_dataset}
\input{5_experiments}
\input{6_conclusion}

\bibliographystyle{IEEEtran}%
\bibliography{main_clean}

\end{document}

%% file: 1_intro.tex
\newcommand{\degree}{\ensuremath{^\circ}}

\begin{table*}[!ht]
\vspace{1.5mm}
\centering
\caption{Our benchmark offers unparalleled precision of depth maps, leveraging varying FOV and high-resolution images, surpassing all existing fisheye datasets. Our training dataset is the first-ever with a vertical stereo setup. }
\resizebox{1.0\textwidth}{!}{%
\begin{tabular}{l|ccccc|cc}
\hline
\multicolumn{1}{c|}{Parameter} & OmniScape & SynWoodScape & Oxford RobotCar & KITTI-360 & WoodScape & \begin{tabular}[c]{@{}c@{}}WideDepth \\ train (ours)\end{tabular} & \begin{tabular}[c]{@{}c@{}}WideDepth \\ benchmark (ours)\end{tabular} \\ \hline
Real/Synthetic & {\color[HTML]{000000} Synthetic} & {\color[HTML]{000000} Synthetic} & Real & {\color[HTML]{000000} Real} & {\color[HTML]{000000} Real} & Real & Real \\
Domain & Outdoor & Outdoor & Outdoor & Outdoor & Outdoor & Outdoor & Indoor \\

Fisheye resolution & 1024$\times$1024 & 1280$\times$966 & 1024$\times$1024 & 1400$\times$1400 & 1280$\times$966 & 1920$\times$1080 & 2048$\times$1152 \\
Fisheye HFOV & 185\degree & 190\degree & 180\degree & 180\degree & 190\degree & 180\degree & 120..195\degree \\
Precision@10m & - & - & $\pm 30$ mm & $\pm 20$ mm & $\pm 20$ mm & $\pm 20$ mm & $\pm 1$ mm \\
Horizontal Stereo & {\color[HTML]{000000} \textcolor{Green}{\cmark}} & {\color[HTML]{000000} \textcolor{Red}{\xmark}} & \textcolor{Red}{\xmark} & {\color[HTML]{000000} \textcolor{Red}{\xmark}} & {\color[HTML]{000000} \textcolor{Red}{\xmark}} & \textcolor{Red}{\xmark} & \textcolor{Green}{\cmark} \\
Vertical Stereo & \textcolor{Red}{\xmark} & \textcolor{Red}{\xmark} & \textcolor{Red}{\xmark} & \textcolor{Red}{\xmark} & \textcolor{Red}{\xmark} & \textcolor{Green}{\cmark} & \textcolor{Green}{\cmark} \\
Pinhole & \textcolor{Red}{\xmark} & \textcolor{Red}{\xmark} & \textcolor{Green}{\cmark} & \textcolor{Green}{\cmark} & \textcolor{Red}{\xmark} & \textcolor{Red}{\xmark} & \textcolor{Green}{\cmark} \\
Dense depth map & \textcolor{Green}{\cmark} & \textcolor{Green}{\cmark} & \textcolor{Red}{\xmark} & \textcolor{Red}{\xmark} & \textcolor{Red}{\xmark} & \textcolor{Red}{\xmark} & \textcolor{Green}{\cmark} \\ \hline
\end{tabular}%
}
\label{tab:dataset_comparison}
\end{table*}

\section{Introduction}
\label{sec:1_intro}

Depth estimation is crucial in autonomous driving, augmented reality (AR), and robotics. Common methods include monocular depth from RGB images, stereo matching, and LiDAR depth completion. However, wide field of view (FOV) depth estimation remains underexplored.

Wide FOV is important for robotics in tight spaces and autonomous vehicles needing 360-degree awareness. Multiple pinhole cameras lead to synchronization and computational issues, while fisheye cameras offer better coverage but pose challenges due to nonlinear geometry and information loss from rectification.

Research on wide FOV and distortion effects remains limited, especially for indoor fisheye depth estimation. Most existing datasets are synthetic or outdoor, lacking real indoor data. To bridge this gap, we introduce a benchmark based on high-precision laser scanning, providing dense point clouds and generated fisheye and pinhole images. As shown in \cref{tab:dataset_comparison}, our benchmark is not only the first in the indoor domain, but also surpasses existing fisheye datasets across multiple parameters.

Notably, stereo matching in the fisheye domain differs significantly from traditional methods. Existing pinhole-based approaches cannot be directly applied, as disparity is interpreted differently in fisheye geometry. However, since usually the primary goal is to obtain metric depth rather than disparity, we developed a method to convert fisheye disparity values to metric depth and vice versa.

The main contributions of our paper are:
\begin{enumerate}
\item We present the first indoor, millimeter-precise fisheye depth benchmark comprising 5K samples across 101 scenes with a wide range of camera parameters. Additionally, we evaluate the performance of state-of-the-art pretrained models on this dataset.

\item We propose a CUDA-accelerated pipeline that generates stereo fisheye RGB, depth, and disparity ground truth from high-resolution LiDAR scans using the Double Sphere model. This enables scalable benchmark creation without storing large panoramas and allows adapting pinhole-trained models to fisheye without retraining.

\item We compile a training dataset of 18K outdoor stereo fisheye pairs with LiDAR depth and demonstrate that additional domain adaptation through fine-tuning of pinhole-based models can further enhance metrics.

\end{enumerate}

%% file: 2_related_work.tex
\section{Related Work}
\label{sec:2_related_work}

\subsection{Depth Estimation Datasets}

NYU-Depth V2 \cite{Silberman:ECCV12NYUDepth} is widely used for indoor depth estimation and instance segmentation, aiding mobile robotics. Captured with Microsoft Kinect, its depth maps have gaps due to structured light limitations. SUN RGB-D \cite{Song2015SUN-RGBD} offers similar data with annotated 2D polygons and 3D cuboids for 10,000 images.

Matterport3D \cite{Chang2017Matterport3D} provides 10,800 panoramic views from 194,400 RGB-D images of 90 large-scale indoor scenes, with camera poses, depth data, and semantic segmentations.

For robotics, IRS \cite{Wang2021IRS} focuses on disparity and surface normal estimation across 103,316 diverse indoor samples. Middlebury \cite{Scharstein2014Middlebury}, despite having a small number of samples, remains a relevant stereo dataset. The Booster \cite{Ramirez2022OBoosterDataset} dataset is also noteworthy, designed for evaluating stereo matching with a focus on challenging glass and mirror surfaces, providing high-resolution pairs and precise ground truth.

\subsection{Fisheye Datasets}

High precision is essential for indoor environments, where manipulation tasks demand sub-centimeter accuracy. Collecting depth data with such precision is challenging, especially for near-field perception, making simulation data a promising alternative. For example, SynWoodScape \cite{Sekkat2022SynWoodScape} and OmniScape \cite{Sekkat2020OmniScape} datasets use simulators, but their scenes are limited to outdoor settings. Additionally, the sim-to-real gap complicates accurate performance evaluation on synthetic data.

The Oxford RobotCar dataset \cite{Maddern2017OxfordRobotCar} provides extensive fisheye camera data from vehicles, but without depth information. KITTI-360 \cite{Liao2021KITTI360} includes fish-eye and pinhole images for outdoor scenes, but its sparse LiDAR depth data only partially covers the wide field of view, limiting its depth estimation utility.

WoodScape \cite{Yogamani2019WoodScape}, inspired by Robert Wood's 1906 fish-eye camera, was among the first to focus on fisheye imagery for tasks like semantic segmentation, object detection, and motion segmentation. However, its lack of depth data remains a significant drawback.

\subsection{Fisheye Projection Models}
\label{sec:2_related_work:proj_models}

Various projection models have been developed to handle depth estimation in super wide FOV applications. The \textbf{equirectangular} projection represents 360° spherical data as a 2D image, mapping latitude and longitude to vertical and horizontal axes. It maintains epipolar line consistency along the vertical axis, offers a continuous full-scene view, and features a regular grid structure, making it well-suited for stereo depth estimation with minimal preprocessing.

The \textbf{cubemap} projection divides the scene into six faces, preserving local detail but introducing discontinuities at face boundaries, which can complicate stereo matching.  A notable example of cubemap-based depth estimation is OmniVidar \cite{xie2023omnividar}.
The \textbf{Cassini} projection was presented in the MODE paper \cite{li2022mode} for depth estimation in horizontal panoramas, maps spherical epipolar lines to sinusoidal curves, simplifying epipolar constraints in such cases. While these projections provide structured representations, their geometric transformations impact depth consistency and processing complexity.

\begin{figure*}[!t]
    \vspace{1.4mm}
    \centering
    \includegraphics[width=1.\textwidth]{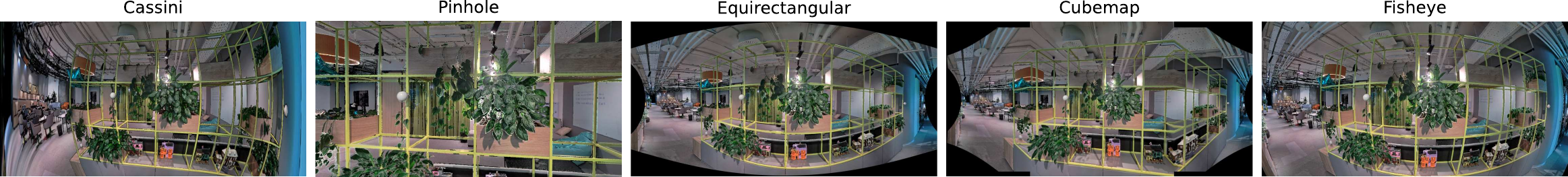}
    \caption{Benchmark scene presented in the considered projections: Cassini, Pinhole, Equirectangular, Cubemap, Fisheye.}
    \label{fig:projections}
\end{figure*}

Other projections, such as \textbf{fisheye} and \textbf{pinhole}, present significant limitations. The \textbf{fisheye} projection captures an ultra-wide FOV but introduces radial distortion, curving epipolar lines and complicating stereo matching. The \textbf{pinhole} model, effective for narrow FOVs, lacks the coverage necessary for wide-angle depth estimation.
All the considered projections were tested on the proposed benchmark as shown in \cref{fig:projections}.

%% file: 3_projection.tex
\section{Methods}
\label{sec:3_projection}

This chapter proposes methods to generate a variety of benchmark data, including different image projections as well as FOV and image distortion. Since the original benchmark data is a set of high-quality colored point clouds, in addition to the images, Depth2Disparity and Disparity2Depth conversions are proposed for equirectangular projection. Equirectangular and cubemap projections were compared for depth estimation on super-wide FOV.

\subsection{Projection and Warping with the DS-Model}
\label{sec:3_projection:warping}
The Double Sphere (DS) camera model proposed by Usenko \textit{et al.} \cite{usenko2018double} is well-suited for cameras with fisheye lenses, offering a closed-form inverse and eliminating the need for computationally expensive trigonometric operations. In the DS model, a 3D point is initially projected onto two unit spheres with their centers offset by $\xi$. The point is then mapped onto the image plane using a pinhole model adjusted by $\alpha/(1 - \alpha)$. The model is defined by the parameter set $\mathbf{i} = [ f_x, f_y, c_x, c_y, \xi, \alpha]^T$.

To warp fisheye images into pinhole or equirectangular views, we first simulate a 3D grid for the target projection and then reproject the points onto the image plane using the Double Sphere model. Camera intrinsics and extrinsics are estimated with the Kalibr toolbox \cite{rehder2016extending,usenko2018double}.

For the set of virtual cameras in the benchmark, the calibrated camera parameters were used as a reference. These cameras are integrated into SensorBox, which is detailed in \cref{sec:4_dataset:hardware}. To capture a diverse set of data, including camera distortion, the intrinsic parameters of the virtual cameras, $\mathbf{i_{virt}}$, were made functionally dependent on the FOV:

\begin{equation}
f_{x,y}^{virt}(\text{FOV}) = f_{x,y} \cdot \left(n \cdot \frac{180^{\circ}}{\text{FOV}}\right),
\end{equation}

\begin{equation}
\xi^{virt}(\text{FOV}) = \xi \cdot \left(1 - m \cdot \frac{\text{FOV}}{180^{\circ}}\right),
\end{equation}

\begin{equation}
\alpha^{virt}(\text{FOV}) = \alpha + m \cdot \left(1 - \alpha\right) \cdot\frac{\text{FOV}}{180^{\circ}},
\end{equation}

The scaling factors $m = 0.2$ and $n = 1.25$ were empirically determined to ensure the preservation of the maximum available density of LiDAR 3D points while maintaining image quality. Specifically, these values were selected to mitigate the occurrence of artifacts, such as blank regions in the projected image. The parameter $f_{x,y}^{virt}$ directly influences the final scale of the projected image. The parameters $\xi^{virt}$ and $\alpha^{virt}$ exhibit a functional dependence on the FOV to explicitly model the increasing impact of distortion as FOV expands.%

\subsection{Disparity-Depth Conversions}
\label{sec:3_projection:disp2depth}

\begin{figure}
    \vspace{-2.mm}
    \centering
    \includegraphics[
    width=0.7\columnwidth,
    keepaspectratio
    ]{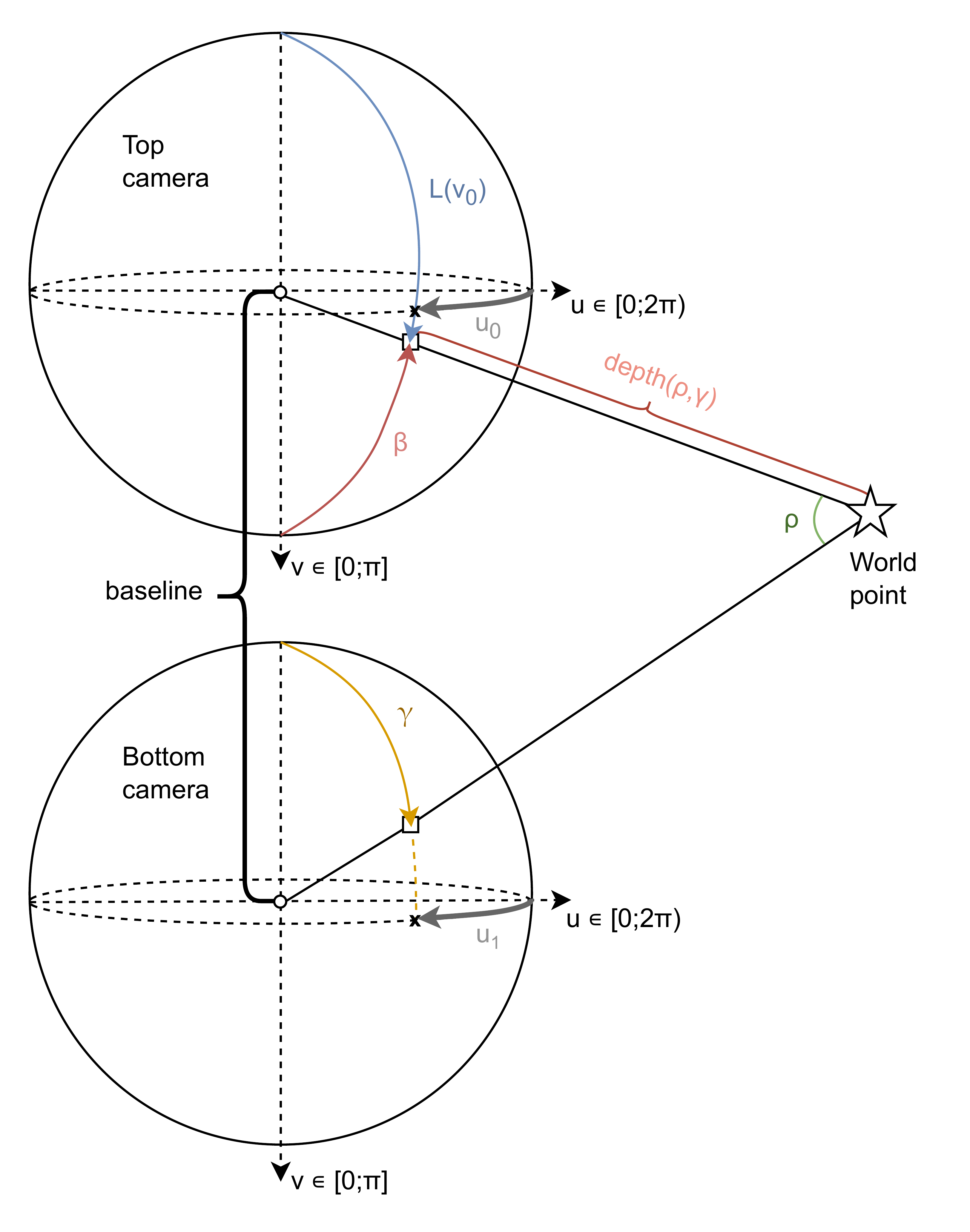}
    \caption{Intuition for the proposed Depth-Disparity Conversion formulations for a vertical stereo pair in spherical projection.}
    \label{fig:vertical_ds}
    \vspace{-1.2mm}
\end{figure}

Stereo Depth Estimation can be performed on a set of virtual vertical stereo pairs in equirectangular projection warped from fisheye. The problem is that Stereo Depth Estimation outputs a disparity map, while initial lidar data can only be naively converted into a depth map. To evaluate the quality of the predicted disparity, we implemented the Disparity2Depth and Depth2Disparity Conversion methods for vertical stereo pairs in equirectangular projection. The geometric intuition for both methods is shown in \cref{fig:vertical_ds}.

\textbf{Disparity2Depth Conversion} on equirectangular projection is based on spherical geometry. Given the disparity map from a stereo pair, the depth for each pixel is calculated by the following steps.

\textbf{1. Latitude Calculation:} The latitude \( L \) of each pixel is determined from its vertical coordinate \( v \) in the image:
\begin{equation}
    L(v) = \frac{v}{H} \cdot \pi,
    \label{eq:L}
\end{equation}

where $L \in [0; \pi]$, \( H \) is the height of the image.

\textbf{2. Disparity Angle:} The disparity angle \( \rho \) for each pixel $\mathbf{u}$ is derived from its disparity value \( disp_{\mathbf{u}} \). $ \rho \in [0; \pi]$ corresponds to the vertex of a 3D point in a spherical triangle:
\begin{equation}
    \rho(\mathbf{u}) = \frac{disp_{\mathbf{u}}}{H} \cdot \pi
    \label{eq:rho}
\end{equation}

\textbf{3. Angles \( \beta \), \( \gamma \):} The angle \( \beta \) is calculated based on the latitude \( L \). The angle \( \gamma \) is dependent on \( \rho \) and \( \beta \) in a spherical triangle:
\begin{equation}
    \beta = \pi - L
    \label{eq:beta}
\end{equation}
\begin{equation}
    \gamma = \pi - \rho - \beta
    \label{eq:gamma}
\end{equation}

\textbf{4. Depth Calculation:} Finally, the $depth(\rho, \gamma)$ for each pixel is determined using the law of sines:
\begin{equation}
depth(\rho, \gamma) = \frac{B \cdot \sin(\gamma)}{\sin(\rho)}
\label{eq:depth}
\end{equation}
where $B$ is a constant factor depending on the stereo camera baseline and setup.

This method ensures an efficient conversion from disparity to depth on equirectangular projections by leveraging the geometry of spherical images. It was validated on Helvipad \cite{zayene2024helvipad}, an omnidirectional dataset with sparse GT depth. While Helvipad is not a fisheye dataset, it is still suitable to evaluate our conversion method. Using CREStereo \cite{li2022crestereo} with our approach, we achieved decent results (MAE = 1.92, RMSE = 3.40), comparable to those reported in the original paper. This demonstrates the robustness of our method, even in challenging scenarios.

\textbf{Depth2Disparity Conversion} on equirectangular projection. The method is predicated on the sine theorem for the aforementioned in \cref{fig:vertical_ds} triangle. The $depth(\rho, \gamma)$ from \cref{eq:depth} is a known value for this method. It will be written $depth_{\mathbf{u}}$. In the following section the transformation to obtain the \( disp_{\mathbf{u}} \) value for each pixel $\mathbf{u}$ is presented:

\begin{equation}
disp_{\mathbf{u}} = \left( \frac{H}{\pi} \right) \cdot \arctan\left( \frac{sin(L_{\mathbf{u}})}{\frac{depth_{\mathbf{u}}}{B} + cos(L_{\mathbf{u}})} \right)
\label{eq:depth2disp}
\end{equation}

This method provides an efficient way to convert depth maps to disparity maps for vertical stereo pairs in equirectangular projection, using the spherical coordinate system.

\textbf{Computational efficiency.} Our implementation of both the Disparity2Depth and Depth2Disparity methods is optimized for computational efficiency through CUDA acceleration, utilizing matrix operations to maximize parallel processing performance. This design enables real-time or near-real-time execution, making it well-suited for high-resolution depth estimation tasks.

\subsection{Projection Model Choice}
\label{sec:3_projection:model_choice}
\textbf{Equirectangular} and \textbf{cubemap} projections were discussed in \cref{sec:2_related_work:proj_models}. These projections offer distinct advantages and challenges in the context of wide FOV depth estimation.

\begin{figure}
    \vspace{1.2mm}
    \centering
    \includegraphics[
    width=0.9\columnwidth,
    keepaspectratio
    ]{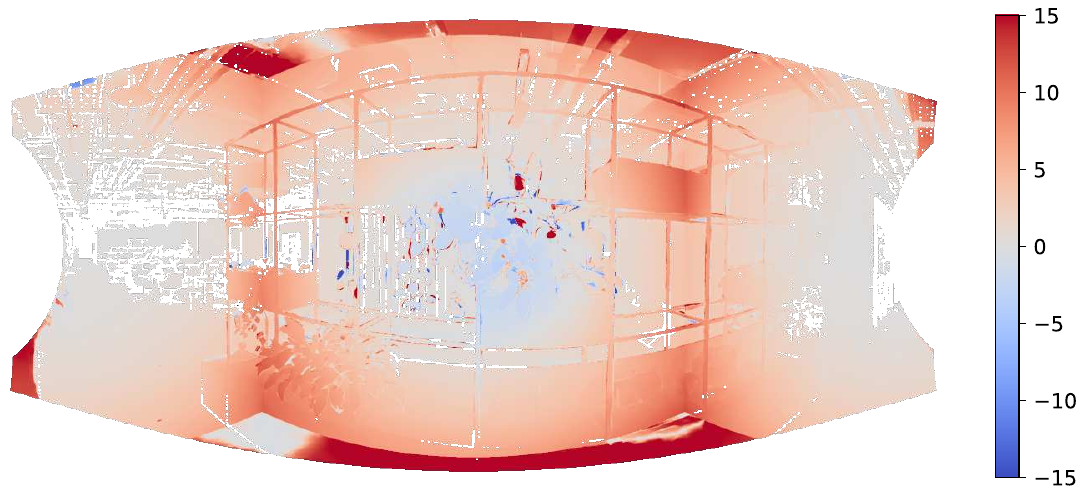}
    \caption{Subtraction of absolute cubemap and equirectangular disparity errors. The prevalence of positive values (red) indicates a larger cubemap error, which grows up at the boundaries.
    }
    \label{fig:cube_error}
\end{figure}

To quantitatively compare these projections, we computed the disparity errors for both models using the reference disparity from our benchmark dataset as GT. Both projections are processed using the same stereo matching model, CREStereo \cite{li2022crestereo}, ensuring that observed differences arise solely from the projection method itself. The difference in absolute errors is visualized in \cref{fig:cube_error}.

A numerical evaluation of disparity errors is provided in \cref{tab:cubemap_vs_equirect}. The equirectangular projection consistently outperforms the cubemap representation across all error metrics, demonstrating lower End-Point Error (EPE) and reduced bad-pixel percentages.

\begin{table}[t]
\centering
\caption{Comparison of projection methods on our WideDepth benchmark. For all metrics the lower the better.}
\resizebox{\columnwidth}{!}{%
\begin{tabular}{l|cccccc}
\hline
Projection & \multicolumn{1}{l}{EPE (px)} & \multicolumn{1}{l}{$Q^{50}_{EPE}$ (\text{px})} & \multicolumn{1}{l}{$Q^{95}_{EPE}$ \text{(px)}} & \multicolumn{1}{l}{bad-1 (\%)} & \multicolumn{1}{l}{bad-2 (\%)} & \multicolumn{1}{l}{bad-3 (\%)} \\ \hline
Cubemap & 4.500 & 3.111 & 13.743 & 78.93 & 62.92 & 51.13 \\
Equirectangular & 1.065 & 0.351 & 3.056 & 28.27 & 10.37 & 5.16 \\ \hline
\end{tabular}%
}
\label{tab:cubemap_vs_equirect}
\end{table}

Overall, while the cubemap projection retains local detail, it suffers from depth consistency issues and increased processing complexity due to face separation and subsequent stitching. The equirectangular projection remains a more structured and effective representation for depth estimation in wide FOV scenarios. Due to these advantages, we use this projection in our experiments.

%% file: 4_dataset.tex
\section{Our Proposed Datasets}
\label{sec:4_dataset}

\begin{figure}
    \vspace{1.4mm}
    \centering
    \includegraphics[
    width=\columnwidth,
    keepaspectratio
    ]{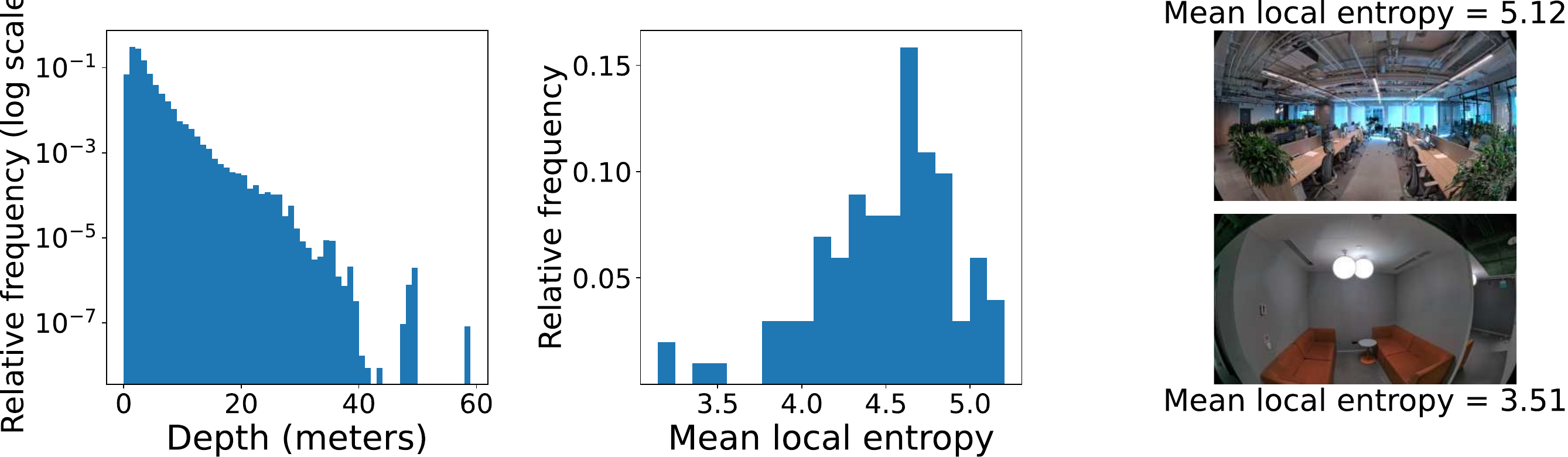}
    \caption{\textbf{Left:} Depth distribution in our benchmark follows typical indoor patterns with a long right tail due to corridors. \textbf{Center:} The distribution of mean local entropy is skewed toward high values, reflecting the abundance of fine structural details and texture complexity. \textbf{Right:} Example of samples with high (top) and low (bottom) entropy; high-entropy images contain small details like plant leaves and ceiling-mounted communication elements.}
    \label{fig:stat_hist}
    \vspace{-1.5mm}
\end{figure}

We introduce two datasets: benchmark and training set. The benchmark, our primary contribution, is smaller but features diverse camera parameters and highly precise GT depth maps. In contrast, our training dataset offers sparse depth maps but includes more samples, enabling us to assess the effect of domain adaptation through fine-tuning.

In this chapter, we detail the requirements and generation processes of both datasets.

\begin{figure*}
  \vspace{1.2mm}
  \centering
  \includegraphics[width=\textwidth]{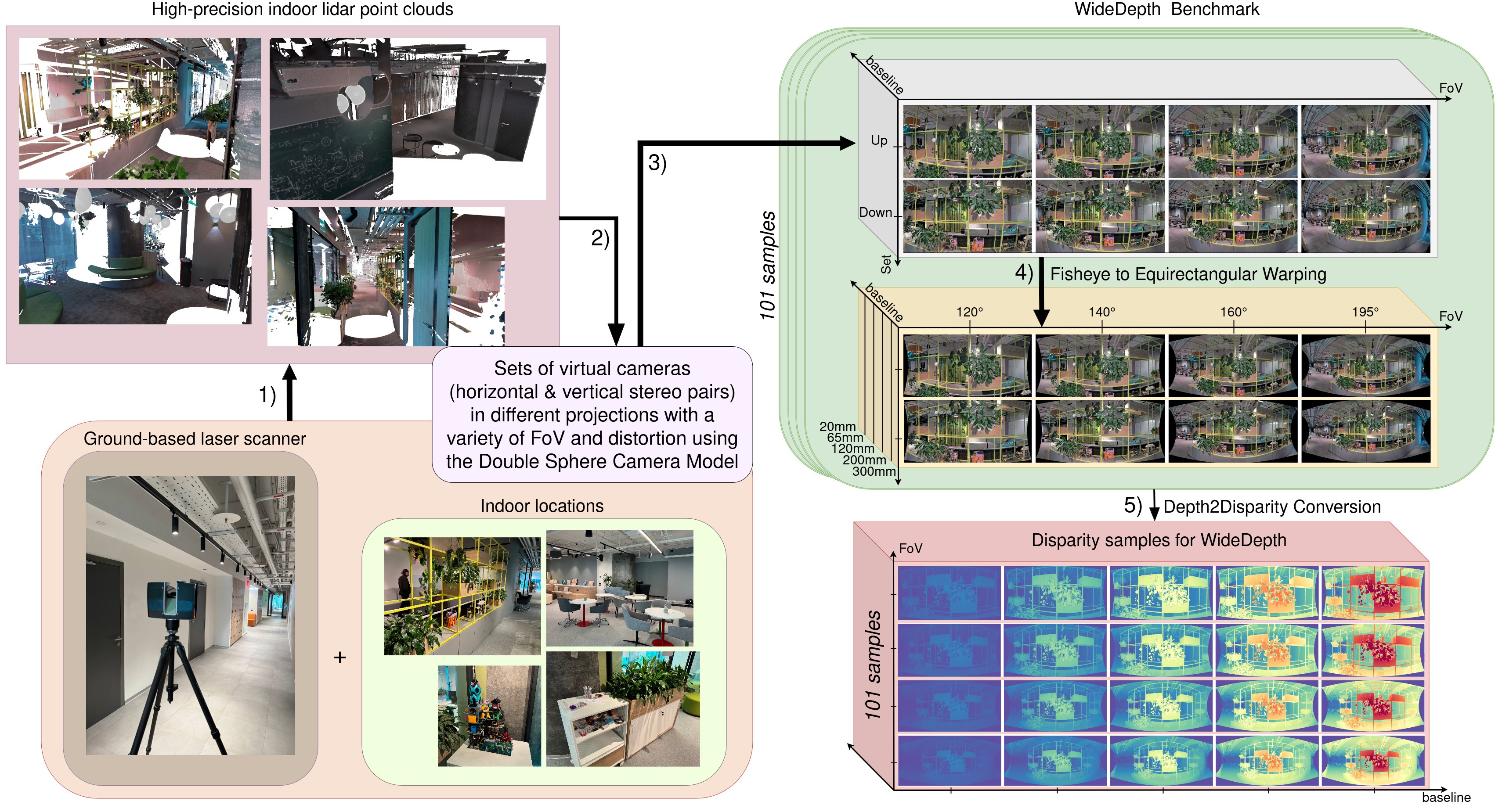}
  \caption{WideDepth: A high-resolution benchmark for fisheye depth estimation in indoor environments. From lidar-generated point clouds (1), we create virtual fisheye cameras with diverse FOVs, distortions, and origins using the Double Sphere Model (2). Cameras capture multiple perspectives and baselines (3), with each fisheye view warped to equirectangular projection to maintain epipolar constraints (4). Finally, depth data is converted to disparity for stereo model evaluation (5). WideDepth includes 101 samples spanning FOVs from 120$^{\circ}$ to 195$^{\circ}$, pushing the limits of mono and stereo depth estimation with superwide-FOV cameras.}
  \label{fig:teaser}
\end{figure*}

\subsection{Benchmark Design}

To evaluate model robustness, a benchmark should include diverse scenes and conditions. Unlike autonomous driving datasets with high-frame-rate vehicle-mounted cameras producing near-duplicate samples, we curate scenes based on fine details, geometric complexity, lighting, and depth estimation use cases. For manipulation, isolated objects are key, while navigation benefits from corridors, offices, and cluttered environments (e.g., kitchens). Captured indoors, our benchmark includes the following lighting categories: natural light, office lamps, and mixed conditions. To enhance diversity, we varied capture heights: 1.65m for AR/humanoid robots, 2.5m for CCTV, and 0.5m for mobile platforms. Higher viewpoints emphasize large planes, while lower ones capture finer details like chair legs and wires.

The key idea is to project a dense, colored point cloud into 2D (see details in \cref{sec:4_dataset:acquisition}), enabling the use of virtual cameras with flexible configurations. For stereo pairs, we utilize a range of baselines tailored to specific applications: from near-field tasks like robotic arm cameras (20 mm) and consumer stereo cameras such as ZED X (65 mm and 120 mm) to setups with extremely large baselines (200 mm and 300 mm). For fisheye cameras, we selected horizontal FOVs of 120, 140, 165, and 195 degrees to evaluate model performance as the angle gradually increases. The combination of five baselines and four angles results in 20 vertical and 20 horizontal stereo pairs per scene. For pinhole cameras, the FOV is fixed at 90 degrees, generating 5 vertical and 5 horizontal pairs per scene based on baseline variations.

\Cref{fig:stat_hist} presents key dataset statistics. The depth range histogram shows a long right tail, which can challenge near-range indoor models. Overall, the primary depth bins are: 0–1 m (6.9\%), 2–5 m (74.7\%), 5–10 m (9.6\%), and 10+ m (1.7\%), aligning with typical indoor datasets. To evaluate scene complexity, we calculated local entropy over 11 neighboring pixels. The majority of samples in our benchmark exhibit high mean local entropy, indicating diverse objects and rich edge details, which pose a greater challenge for depth estimation methods.

We generate stereo fisheye pairs directly from merged LiDAR scans, avoiding the need for panoramic intermediates. With the Double Sphere camera model and a CUDA-accelerated pipeline, our method efficiently synthesizes fisheye images with arbitrary baselines and focal lengths, preserving realistic occlusions and geometric fidelity. This design provides broad coverage of configurations and scalable image generation for diverse benchmarking scenarios.

\subsection{Hardware}
\label{sec:4_dataset:hardware}
To obtain a high-precision colored point cloud for benchmark, we used a ground-based laser scanner with a range error of 1-2 mm and a resolution up to 165 MP. Each scan generates about 20 million points over a 360° view but requires 10 minutes per scan, limiting large-scale data collection. For train dataset we developed the portable SensorBox, integrating pair of ZED X One fisheye cameras (30 Hz) in vertical stereo setup and two Livox Mid 360 LiDARs (10 Hz) positioned such that their fields
of view partially overlap, maximizing coverage of the scene.
Further details, including LiDAR specifications and SensorBox setup, can be found in the accompanying video clip.

\subsection{Datasets Acquisition}
\label{sec:4_dataset:acquisition}

\textbf{Benchmark.} We performed laser scanning in the office space. The distance between scanning points varied from 1.5 to 5 meters, with closer spacing in rooms containing furniture or small objects and wider spacing in long corridors. Spherical markers, positioned around the scanner, facilitated alignment of neighboring point clouds. To generate 2D depth maps, point clouds were sampled and projected based on the virtual camera's direction, baseline, and FOV. This pipeline can be seen on the \cref{fig:teaser}. Stereo pairs were created by shifting the camera parallel to the scan center, which can introduce occlusions or empty regions, especially with wider baselines. To address this, we used a bundle of three overlapping scans, allowing occlusions in the \textit{central} scan to be filled with points from \textit{adjacent} scans. This method was applied to all but the first and last scans, resulting in a total of 101 scenes. Windows and mirrors are handled through semi-manual point cloud cleaning. To maintain indoor depth distribution, we clip the depth of outdoor objects visible through windows to a closer value. Also, reflective surfaces are masked with zero depth values.

\textbf{Train dataset.} To prevent overfitting to benchmark indoor environment, our train dataset was captured entirely outdoors, handheld, across city areas, parks, and streets during daylight under cloudy conditions. Stereo pairs were rectified and warped to an equirectangular projection as described in \cref{sec:3_projection:warping}. Point clouds from two hardware-synced LiDARs were merged into a single cloud and projected onto the upper camera frame using the DS camera model. For the training set, we prioritized dataset size over GT label precision and density as a trade-off. Synchronization between LiDARs and the camera was achieved through timestamp alignment within the ROS framework. In addition to \textit{depth} GT, we generated \textit{disparity} GT using our proposed Depth2Disparity conversion method described in \cref{sec:3_projection:disp2depth}. After projecting the point clouds into the camera frame, we observed a mean depth point density of 8.1\% across the dataset.

%% file: 5_experiments.tex
\section{Experiments}
\label{sec:5_experiments}
In this chapter, we evaluate SOTA models on our benchmark, analyzing how increasing FOV impacts monocular depth and depth completion models. We expect monocular models to degrade more noticeably than depth completion models, which rely on depth guidance. We also evaluate stereo models across varying FOV-baseline combinations and fine-tune a pretrained stereo model on our fisheye dataset to assess domain adaptation effects.

\subsection{Monocular and Depth Completion}

\begin{figure}[t]
    \vspace{1.4mm}
    \centering
    \includegraphics[
    width=0.8\columnwidth,
    keepaspectratio
    ]{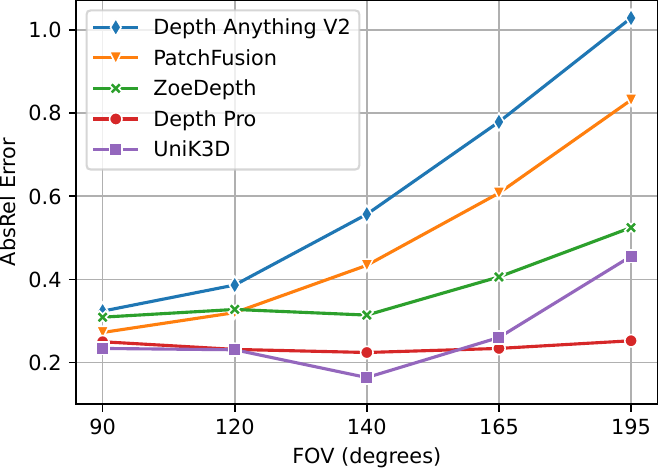}
    \caption{All observed monocular depth models perform well on pinhole images, but most degrade with a wider field of view. This deterioration at higher FOV highlights the unresolved performance issue, making our benchmark highly valuable.
    }
    \label{fig:line_mono}
    \vspace{-4.2mm}
\end{figure}

\textbf{Monocular Depth.} For comparison, we selected SOTA off-the-shelf models that offer acceptable inference times and memory consumption for practical applications, with an additional key criterion being their ability to produce metric depth. Based on these requirements, we chose Depth Anything V2-Large \cite{Yang2024DepthAV2}, PatchFusion \cite{Li2023PatchFusionAE}, ZoeDepth \cite{Bhat2023ZoeDepthZT}, Depth Pro \cite{bochkovskii2024depthpro} and UniK3D-Large \cite{piccinelli2025unik3d} models.

Following the standard depth estimation metrics summarized in \cite{masoumian2022mde_survey}, we analyzed how quality changes as the FOV expanded from 120 to 195 degrees, the narrowest to widest fisheye in our dataset. \Cref{tab:mono_dc_metric_delta} shows that all monocular models degrade at wider FOVs, but to varying extents. We further examined model performance at intermediate FOV values. As depicted in \cref{fig:line_mono}, nearly all models exhibit gradual performance degradation as FOV increases. Since UniK3D was explicitly designed to be camera-universal, it achieves its highest accuracy at moderate fisheye ranges (120–140$^{\circ}$) and surpasses other tested monocular models in this regime. This makes it the most reliable choice for applications targeting wide yet not extreme FOVs. However, even this model shows decline at extreme FoVs ($\ge 165^{\circ}$) when operated in camera-free mode without intrinsics.

This supports our initial hypothesis that wider angles make it more challenging for models to achieve metric depth due to domain shift and distorted geometry.

\begin{table}
\vspace{1.4mm}
\centering
\caption{Percentage change in metrics for FOV 195 compared to FOV 120. Green indicates improved results with a wider angle, while red - decline in performance at a wider angle.}
\label{tab:mono_dc_metric_delta}
\resizebox{\columnwidth}{!}{%
\begin{tabular}{l|lll|lll}
\hline
Model & AbsRel $\downarrow$ & MAE $\downarrow$ & RMSE $\downarrow$ &
$\delta_{1.25}$ $\uparrow$ & $\delta_{1.25^2}$ $\uparrow$ & $\delta_{1.25^3}$ $\uparrow$ \\ \hline
NLSPN \cite{park2020nlspn} & {\color[HTML]{9A0000} +5.92\%} & {\color[HTML]{009901} -1.26\%} & {\color[HTML]{009901} -3.05\%} & {\color[HTML]{9A0000} -0.03\%} & {\color[HTML]{009901} +0.04\%} & {\color[HTML]{009901} +0.04\%} \\
CompletionFormer \cite{Zhang2023CompletionFormerDC} & {\color[HTML]{9A0000} +6.73\%} & {\color[HTML]{009901} -2.05\%} & {\color[HTML]{9A0000} +1.97\%} & {\color[HTML]{9A0000} -0.01\%} & {\color[HTML]{009901} +0.05\%} & {\color[HTML]{009901} +0.04\%} \\
CostDCNet \cite{Kam2022CostDCNetCV} & {\color[HTML]{9A0000} +38.5\%} & {\color[HTML]{009901} -13.9\%} & {\color[HTML]{9A0000} +0.67\%} & {\color[HTML]{009901} +0.06\%} & {\color[HTML]{9A0000} -0.07\%} & {\color[HTML]{9A0000} -0.16\%} \\ \hline
DepthAnythingV2 \cite{Yang2024DepthAV2} & {\color[HTML]{9A0000} +166\%} & {\color[HTML]{9A0000} +126\%} & {\color[HTML]{9A0000} +106\%} & {\color[HTML]{9A0000} -86.8\%} & {\color[HTML]{9A0000} -81.5\%} & {\color[HTML]{9A0000} -53.9\%} \\
PatchFusion \cite{Li2023PatchFusionAE} & {\color[HTML]{9A0000} +160\%} & {\color[HTML]{9A0000} +123\%} & {\color[HTML]{9A0000} +91.3\%} & {\color[HTML]{9A0000} -87.4\%} & {\color[HTML]{9A0000} -68\%} & {\color[HTML]{9A0000} -27.9\%} \\
UniK3D \cite{piccinelli2025unik3d} & {\color[HTML]{9A0000} +97.3\%} & {\color[HTML]{9A0000} +147.2\%} & {\color[HTML]{9A0000} +119.6\%} & {\color[HTML]{9A0000} -73.7\%} & {\color[HTML]{9A0000} -56.2\%} & {\color[HTML]{9A0000} -38.9\%} \\
ZoeDepth \cite{Bhat2023ZoeDepthZT} & {\color[HTML]{9A0000} +60.1\%} & {\color[HTML]{9A0000} +47.3\%} & {\color[HTML]{9A0000} +21.3\%} & {\color[HTML]{9A0000} -51.7\%} & {\color[HTML]{9A0000} -25\%} & {\color[HTML]{9A0000} -6.5\%} \\
DepthPro \cite{bochkovskii2024depthpro} & {\color[HTML]{9A0000} +9.0\%} & {\color[HTML]{9A0000} +3.3\%} & {\color[HTML]{9A0000} +8.3\%} & {\color[HTML]{9A0000} -4.1\%} & {\color[HTML]{9A0000} -4.5\%} & {\color[HTML]{9A0000} -2.31\%} \\ \hline
\end{tabular}%
}
\end{table}

\textbf{Depth Completion.} Among state-of-the-art models for depth completion, we selected two larger models, NLSPN \cite{park2020nlspn} and CompletionFormer \cite{Zhang2023CompletionFormerDC}, as well as a lighter model, CostDCNet \cite{Kam2022CostDCNetCV}, which offers high quality with low latency. All three models were trained on the same NYUv2 \cite{Silberman:ECCV12NYUDepth} dataset. As shown in \cref{tab:mono_dc_metric_delta}, although all models experience a decrease in quality according to the AbsRel metric, certain metrics actually improve in the larger models. This may be attributed to complex models — those incorporating mechanisms like attention and higher capacity — being better equipped to interpret scenes at wide viewing angles.

In summary, even robust, large-scale models like Depth Anything V2-Large degrade in performance on fisheye images as FOV increases.
As expected, depth completion models are less affected by wide angles but still show significant declines in the AbsRel metric, one of the most important indicators, even among high-performing SOTA models with attention mechanisms.

\begin{figure*}
    \vspace{1.4mm}
    \centering
    \includegraphics[width=.84\textwidth]{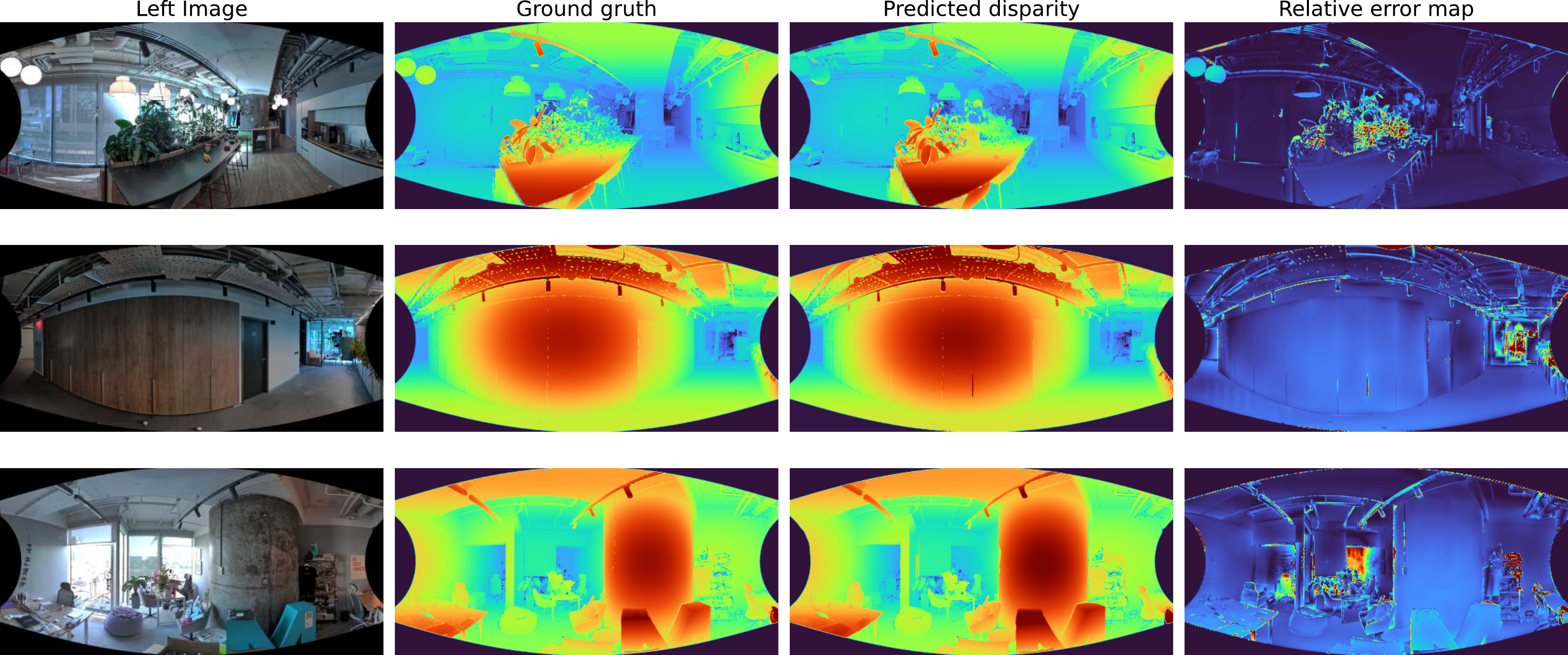}
    \caption{Qualitative results for stereo with FOV 195 using the StereoBase model. The model shows no degradation from geometric distortion, demonstrating the success of our approach in adapting pinhole models to fisheye data.}
    \label{fig:stereo_qualitative}
\end{figure*}

\subsection{Stereo Matching and Fine-tuning}

Stereo models vary widely in size, so we selected models from different weight categories, including FADNet++ \cite{Wang2021FADNetRA}, CREStereo \cite{li2022crestereo}, BGNet \cite{xu2021bgnet}, IGEV \cite{xu2023igev}, GMStereo \cite{xu2023unimatch}, and StereoBase \cite{guo2024openstereo}. All models received input data processed as described in \cref{sec:3_projection}, including equirectangular projection, cropping of empty areas, and a 90-degree counter-clockwise rotation. Evaluation followed the commonly adopted stereo metrics summarized in \cite{tosi2025stereo_survey}. We also measured each model’s latency in half precision on an NVIDIA GeForce RTX 3080 Ti at 1024 × 512 resolution.

\begin{figure}
    \vspace{-0.5mm}
    \centering
    \includegraphics[
    width=0.83\columnwidth,
    keepaspectratio
    ]{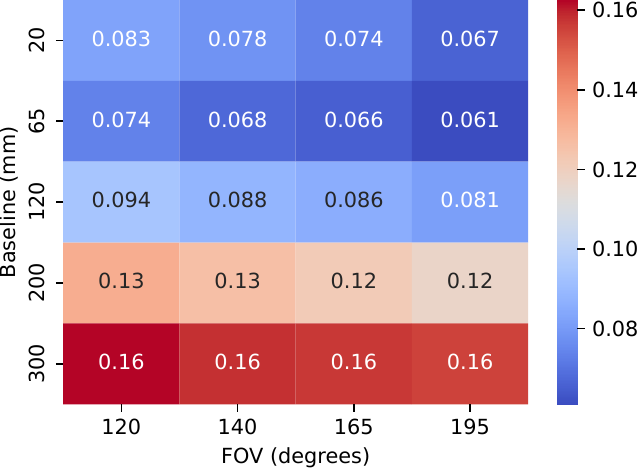}
    \caption{Impact of baseline and FOV variations on RelEPE (lower is better) using StereoBase model. Results indicate much greater sensitivity to unfamiliar baselines than high FOVs, demonstrating our approach's effectiveness for fisheye stereo.}
    \label{fig:heatmap}
    \vspace{-2.5mm}
\end{figure}

Quantitative results in \cref{tab:stereo_improve} show that model rankings closely match pinhole benchmarks, suggesting that fisheye stereo performance can be inferred from pinhole-based ratings. For wide-angle applications, StereoBase proves ideal, though at the cost of high latency. In \cref{fig:stereo_qualitative} this model demonstrates consistent performance with our method, even in challenging cases. The relative error map highlights common stereo matching errors (e.g., on translucent and reflective surfaces) without fisheye-specific issues, showcasing the robustness of our adaptation approach.

We also analyzed the impact of baseline and FOV on StereoBase, finding that baseline variations affect performance more significantly than FOV. A 65 mm baseline yields optimal metrics, while larger baselines degrade quality due to disparity distribution shift. Conversely, higher FOV consistently enhances performance, reinforcing the importance of wide angles for capturing broader indoor environments.

\textbf{Fine-tuning.} We selected BGNet, a lightweight model optimized for real-time inference on embedded devices (e.g., Jetson Orin), making it well-suited for practical use. A SceneFlow-pretrained model was fine-tuned on our WideDepth training set and evaluated on the WideDepth benchmark under the same conditions as other pretrained models. Fine-tuning was conducted for 15 epochs with batch size 8, half precision, and the AdaBelief \cite{Zhuang2020AdaBeliefOA} optimizer. A one-cycle scheduler \cite{Smith2018SuperconvergenceVF} adjusted the learning rate, peaking at $5 \cdot 10^{-5}$ with a warmup over 10\% of iterations. To increase data diversity, we applied asymmetric chromatic augmentation with 50\% probability.

\Cref{tab:stereo_improve} shows substantial performance gains after fine-tuning on our dataset, matching or surpassing heavier models. This underscores that while robust fisheye performance can be achieved with pinhole-trained models, as we proposed, dedicated domain adaptation to fisheye data can further improves results.

\begin{table}
\vspace{-1.2mm}
\centering
\caption{Stereo model performance at 195$^\circ$ FOV and 65 mm baseline; lower values are better. Fine-tuning lightweight BGNet on our train dataset boosts it to match heavier SOTA models, showing the value of domain adaptation}
\resizebox{\columnwidth}{!}{%
\begin{tabular}{l|c|cccccc}
\hline
Model & \begin{tabular}[c]{@{}c@{}}Latency\\ (ms)\end{tabular} & \begin{tabular}[c]{@{}c@{}}EPE\\ (px)\end{tabular} & \begin{tabular}[c]{@{}c@{}}$Q_{EPE}^{50}$\\ (px)\end{tabular} & \begin{tabular}[c]{@{}c@{}}$Q_{EPE}^{95}$\\ (px)\end{tabular} & bad-1 (\%) & bad-2 (\%) & bad-3 (\%) \\ \hline
FADNet++ \cite{Wang2021FADNetRA}& 40 & 2.089 & 0.643 & 6.373 & 29.63 & 15.59 & 10.46 \\
GMStereo \cite{xu2023unimatch}& 186 & 1.950 & 0.700 & 6.495 & 30.06 & 15.16 & 10.31 \\
IGEV \cite{xu2023igev}& 552 & \textbf{1.605} & 0.533 & \textbf{4.526} & 24.87 & 12.58 & \textbf{8.37} \\
StereoBase \cite{guo2024openstereo}& 665 & 1.612 & 0.529 & 4.554 & \textbf{24.75} & \textbf{12.49} & \textbf{8.37} \\
CREStereo \cite{li2022crestereo}& 284 & 1.759 & \textbf{0.528} & 6.238 & 24.78 & 13.02 & 9.02 \\
BGNet \cite{xu2021bgnet}& 19 & 3.411 & 1.410 & 13.595 & 53.30 & 33.32 & 24.34 \\ \hline
\begin{tabular}[c]{@{}l@{}}BGNet \cite{xu2021bgnet}\\ (fine-tuned)\end{tabular} & 19 & {\color[HTML]{000000} \begin{tabular}[c]{@{}c@{}}1.767\\ \textcolor{ForestGreen}{(-48\%)}\end{tabular}} & \begin{tabular}[c]{@{}c@{}}0.606\\ \textcolor{ForestGreen}{(-57\%)}\end{tabular} & \begin{tabular}[c]{@{}c@{}}5.736\\ \textcolor{ForestGreen}{(-58\%)}
\end{tabular} & \begin{tabular}[c]{@{}c@{}}27.63\\ \textcolor{ForestGreen}{(-48\%)}\end{tabular} & \begin{tabular}[c]{@{}c@{}}13.76\\ \textcolor{ForestGreen}{(-59\%)}\end{tabular} & \begin{tabular}[c]{@{}c@{}}9.29\\ \textcolor{ForestGreen}{(-62\%)}\end{tabular} \\ \hline
\end{tabular}%
}
\label{tab:stereo_improve}
\vspace{-1.2mm}
\end{table}

%% file: 6_conclusion.tex
\section{Limitations and Future work}
\label{sec:6_limitations}

LiDAR scanning struggles to accurately capture transparent and reflective surfaces, so we masked these areas in our benchmark. However, these challenging cases present valuable research opportunities, and we aim to address depth estimation in such conditions in future work. In addition, we plan to expand our benchmark beyond depth estimation by incorporating other vision tasks, such as semantic segmentation, to enhance its applicability.
Regarding the experiments, our GPU constraints restricted fine-tuning to lightweight architectures, leaving the effect of domain adaptation on larger models with distorted geometry uncertain.

\section{Conclusion}
\label{sec:7_conclusion}

In this article, we introduced WideDepth, the first indoor fisheye depth benchmark with 101 scenes and over 5K stereo pairs featuring diverse fields of view, baselines, and high-precision depth and disparity maps. Additionally, we provided an outdoor stereo fisheye dataset containing 18K LiDAR-labeled samples, which has demonstrated strong effectiveness for model adaptation to fisheye geometry.

Since existing disparity-to-depth methods do not directly apply to fisheye stereo, we proposed a conversion method tailored for fisheye cases. To leverage a broad range of existing stereo models, we introduced an approach that enables pinhole-trained models to be used on fisheye images without architectural changes. Finally, we evaluated 14 SOTA models across various tasks on our benchmark, including monocular depth estimation, stereo matching, and depth completion.

We believe WideDepth will significantly advance fisheye depth estimation research, bridging the gap with pinhole-based methods.

\section{Acknowledgement}
The work of Ilya Makarov on Section 2 was supported by a grant, provided by the Ministry of Economic Development of the Russian Federation (agreement dated June 20, 2025 No. 139-15-2025-011, identifier 000000C313925P4G0002).